\documentclass[runningheads]{llncs}

\usepackage{tikz}
\usepackage{multirow}
\usepackage{graphicx}
\usepackage{comment}
\usepackage{amsmath,amssymb} 
\usepackage{color}

\usepackage{epsfig}
\usepackage{graphicx}
\usepackage{mathrsfs}
\usepackage{url}
\usepackage{enumitem}
\usepackage{algpseudocode}
\usepackage{dsfont}
\usepackage{multirow}
\usepackage{booktabs}
\usepackage[pagebackref=false,breaklinks=true,letterpaper=true,colorlinks,urlcolor=blue,citecolor=blue,linkcolor=blue,bookmarks=false]{hyperref}

\begin{document}
\pagestyle{headings}
\mainmatter
\def\ECCVSubNumber{4179}  

\title{Rethinking Image Inpainting via a Mutual Encoder-Decoder with Feature Equalizations} 

\titlerunning{Deep Image Inpainting}
%
\author{Hongyu Liu$^1$ \and
Bin Jiang$^{1\star}$ \and
Yibing Song$^2$\thanks{B. Jiang and Y. Song are the corresponding authors. This work is done partially when H. Liu is an intern in Tencent AI Lab. The results and code are available on \url{https://github.com/KumapowerLIU/Rethinking-Inpainting-MEDFE}.} \and
Wei Huang$^1$ \and
Chao Yang$^1$}
\authorrunning{Liu et al.}
%

\institute{College of Computer Science and Electronic Engineering, Hunan University, Changsha, China\\
\email{\{kumapower,jiangbin,hwei,yangchaoedu\}@hnu.edu.cn }
 \and
Tencent AI Lab\\
\email{yibingsong.cv@gmail.com}}
\maketitle

\begin{abstract}
Deep encoder-decoder based CNNs have advanced image inpainting methods for hole filling. While existing methods recover structures and textures step-by-step in the hole regions, they typically use two encoder-decoders for separate recovery. The CNN features of each encoder are learned to capture either missing structures or textures without considering them as a whole. The insufficient utilization of these encoder features limit the performance of recovering both structures and textures. In this paper, we propose a mutual encoder-decoder CNN for joint recovery of both. We use CNN features from the deep and shallow layers of the encoder to represent structures and textures of an input image, respectively. The deep layer features are sent to a structure branch and the shallow layer features are sent to a texture branch. In each branch, we fill holes in multiple scales of the CNN features. The filled CNN features from both branches are concatenated and then equalized. During feature equalization, we reweigh channel attentions first and propose a bilateral propagation activation function to enable spatial equalization. To this end, the filled CNN features of structure and texture mutually benefit each other to represent image content at all feature levels. We use the equalized feature to supplement decoder features for output image generation through skip connections. Experiments on the benchmark datasets show the proposed method is effective to recover structures and textures and performs favorably against state-of-the-art approaches.
\end{abstract}

\section{Introduction}

There is a need to recover missing contents in corrupted images for visual aesthetics improvement. Deep neural networks have advanced image inpainting by introducing semantic guidance to fill hole regions. Different from the traditional methods~\cite{Efros-sig01-quilting1,Barnes-sig09-patchmatch3,Efros-ICECCS01-synthesis2,bertalmio-SIG2000-imageinpainting18} that propagate uncorrupted image contents to the hole regions via patch-based image matching, deep inpainting methods~\cite{Pathak-cvpr16-featinpaint7,Iizuka-sig17-lgmatch4} utilize CNN features in different levels (i.e., from low-level features to high-level semantics) to produce more meaningful and globally consistent results.

\def\swone{0.9\linewidth}
\renewcommand{\tabcolsep}{1pt}
\def\swfive{0.19\linewidth}
\def\swfivep{0.08\linewidth}
\begin{figure}[t]
\begin{center}
\begin{tabular}{ccccc}
\includegraphics[width=\swfive]{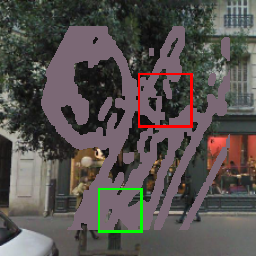}&
\includegraphics[width=\swfive]{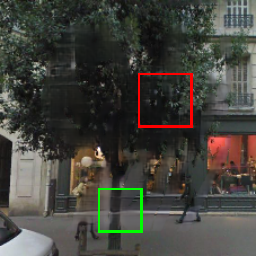}&
\includegraphics[width=\swfive]{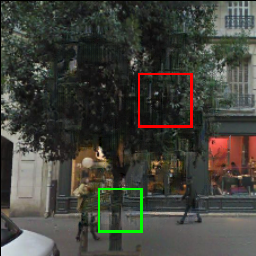}&
\includegraphics[width=\swfive]{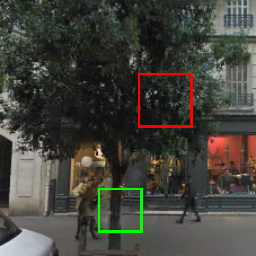}&
\includegraphics[width=\swfive]{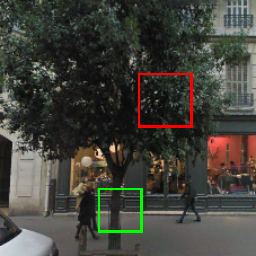}\\
\includegraphics[width=\swfivep]{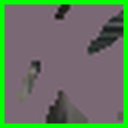} \includegraphics[width=\swfivep]{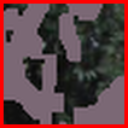}&
\includegraphics[width=\swfivep]{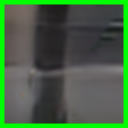} \includegraphics[width=\swfivep]{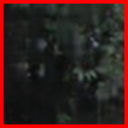}&
\includegraphics[width=\swfivep]{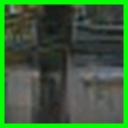} \includegraphics[width=\swfivep]{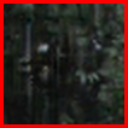}&
\includegraphics[width=\swfivep]{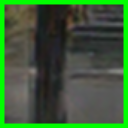} \includegraphics[width=\swfivep]{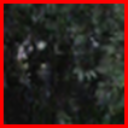}&
\includegraphics[width=\swfivep]{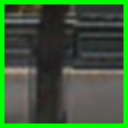} \includegraphics[width=\swfivep]{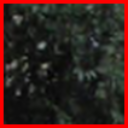}\\
(a) Input &(b) GC~\cite{yu-2019iccv-freeform38}&(c) CSA~\cite{liu-2019iccv-coherent44}&(d) Ours&(e) GT\\
\end{tabular}
\end{center}
\caption{Visual comparison on the Paris StreetView dataset~\cite{doersch-2015acm-makesparis39}. GT is the ground truth image. The proposed inpainting method is effective to reduce blur and artifacts within and around the hole regions, which are brought by inconsistent structure and texture features.}
\label{fig:intro}
\vspace{-2mm}
\end{figure}

The encoder-decoder architecture is prevalent in existing deep inpainting methods~\cite{Iizuka-sig17-lgmatch4,Li-cvpr17-facecomplete5,Yeh-arxiv1607-spinpaint6,Pathak-cvpr16-featinpaint7}. However, a direct utilization of the end-to-end training and prediction processes generate limited results. This is due to the challenging factor that the hole region is completely empty. Without sufficient image guidance, an encoder-decoder is not able to reconstruct the whole missing content. An alternative is to use two encoder-decoders to separately learn missing structures and textures in a step-by-step manner. These two-stage methods~\cite{song-2018arxiv-spg32,nazeri-2019ICCVW-edgeconnect46,ren-2019iccv-structureflow47,Yu-cvpr18-attentioninpainting8,Song-eccv18-iftinpainting10,yu-2019iccv-freeform38,liu-2019iccv-coherent44} typically generate an intermediate image with recovered structures in the first stage (i.e., encoder-decoder), and send this image to the second stage for texture generation. Although structures and textures are produced on the output image, their appearances are not consistent. Fig.~\ref{fig:intro} shows an example, the inconsistent structures and textures within hole regions produce blur and artifacts as shown in (b) and (c). Meanwhile, the recovered contents are not coherent to the uncorrupted contents around the hole boundaries (e.g., the leaves). This limitation is because of the independent learning of CNN features representing structures and textures. In practice, the structures and textures correlate with each other to formulate the image contents. Without considering their coherence, existing methods are not able to produce visually pleasing results.

In this work, we propose a mutual encoder-decoder to jointly learn CNN features representing structures and textures. The features from the deep layers of the encoder contain structure semantics while the features from the shallow layers contain texture details. The hole regions of these two features are filled via two separate branches. In the CNN feature space, we use a multi-scale filling block within each branch for hole filling. Each block consists of 3 partial convolution streams with progressively increased kernel sizes. After hole filling in these two features, we propose a feature equalization method to ensure the structure and texture features consistent with each other. Meanwhile, the equalized features are coherent with the features of uncorrupted image content around the hole boundaries. The proposed feature equalization consists of channel reweighing and bilateral propagation. We concatenate two features first and perform channel reweighing via attention exploration~\cite{hu-2018CVPR-squeeze55}. The attentions across two features are set to be consistent after channel equalization. Then, we propose a bilateral propagation activation function to equalize the feature consistency in the whole feature maps. This activation function uses elements on the global feature maps to propagate channel consistency (i.e., feature coherence across the hole boundaries), while using elements within local neighboring regions to maintain channel similarities (i.e., feature consistency within the hole). To this end, we fuse the texture and structure features together to reduce inconsistency in the CNN feature maps. The equalized features then supplement the decoder features in all the feature levels via encoder-decoder skip connections. The feature consistency is then reflected in the reconstructed output image, where the blur and artifacts are effectively removed around the hole regions as shown in Fig.~\ref{fig:intro}(d). Experiments on the benchmark datasets show that the proposed method performs favorably against state-of-the-art approaches.

We summarize the contributions of this work as follows:
\begin{itemize}[noitemsep,nolistsep]
  \item We propose a mutual encoder-decoder network for image inpainting. The CNN features from the shallow layer are learned to represent textures and the features from deep layers represent structures.
  \item We propose a feature equalization method to make structure and texture features consistent with each other. We first reweigh channels after feature concatenation and propose a bilateral propagation activation function to make the whole feature consistent.
  \item Extensive experiments on the benchmark datasets have shown the effectiveness of the proposed inpainting method in removing blur and artifacts caused by inconsistent structure and texture features. The proposed method performs favorably against state-of-the-art inpainting approaches.
\end{itemize}

\section{Related Works}
{\flushleft \bf Empirical Image Inpainting}.
The empirical image inpainting methods~\cite{bertalmio-SIG2000-imageinpainting18,levin-2003-learning49,ballester-2001tio-filling50} based on diffusion techniques propagate the neighborhood appearances to the missing regions. However, they only consider surrounding pixels of missing regions, which can only deal with small holes in background inpainting tasks and may fail to generate meaningful structures. In contrast, methods~\cite{criminisi-2004tip-region20,darabi-2012tog-imagemelding28,song2017stylizing,Barnes-sig09-patchmatch3,xu-2010-image25}  based on patch match fill missing regions by transferring similar and relevant patches from the remaining image region to the hole region. Although empirical methods perform well to handle small holes on the background inpainting task, they are not able to generate semantically meaningful content. When the hole region is large, these methods suffer from a lack of semantic guidance.

{\flushleft \bf Deep Image Inpainting}.
Image inpainting based on deep learning typically involves the generative adversarial network~\cite{goodfellow-2014nips-generative40} to supplement visual perceptual guidance for hole filling. Pathak et al.~\cite{Pathak-cvpr16-featinpaint7} first bring adversarial training~\cite{goodfellow-2014nips-generative40} to inpainting and demonstrate semantic hole-filling. Iizuka et al.~\cite{Iizuka-sig17-lgmatch4} propose local and global discriminators, assisted by dilated convolution~\cite{yu-2015-multidilated58} to improve the inpainting quality. Nazeri et al.~\cite{nazeri-2019ICCVW-edgeconnect46} propose EdgeConnect that predicts salient edges for inpainting guidance. Song et al.~\cite{song-2018arxiv-spg32} utilize a segmentation prediction network to generate segmentation guidance for detail refinement around the hole region. Xiong et al.~\cite{xiong-2019CVPR-foreground60} present foreground-aware inpainting, which involves three stages, i.e., contour detection, contour completion and image completion, for the disentanglement of structure inference and content hallucination. Ren et al.~\cite{ren-2019iccv-structureflow47} introduce structure-aware network, which splits the inpainting task into two parts: structure reconstruction and texture generation. It uses appearance flow to sample features from contextual regions. Yan et al.~\cite{Yan-eccv18-shiftnet9} speculate the relationship between the contextual regions in the encoder layer and the associated hole region in the decoder layer for better predictions. Yu et al.~\cite{Yu-cvpr18-attentioninpainting8} and Song et al.~\cite{Song-eccv18-iftinpainting10} search for a collection of background patches with the highest similarity to the generated contents in the first stage prediction. Liu et al.~\cite{liu-2018eccv-particalcov37} address this inpainting task via exploiting the partial convolutional layer and mask-update operation. Following the~\cite{liu-2018eccv-particalcov37}, Yu et al.~\cite{yu-2019iccv-freeform38} present gate convolution that learns a dynamic mask-update mechanism and combines with SN-PatchGAN discriminator to achieve better predictions.  Liu et al.~\cite{liu-2019iccv-coherent44} propose coherent semantic attention, which considers the feature coherency of hole regions to guarantee the pixel continuity in image level. Wang et al.~\cite{wang-2018NIPS-imageINPAINTING59} propose a generative multi-column convolutional neural network (GMCNN) that uses varied receptive fields in branches. Different from existing deep inpainting methods, our method produces CNN features to consistently represent structures and textures to reduce blurry and artifacts around the hole region.

\section{Proposed Algorithm}
\begin{figure}[t]
\begin{center}
\begin{tabular}{c}
\includegraphics[width=1.0\linewidth]{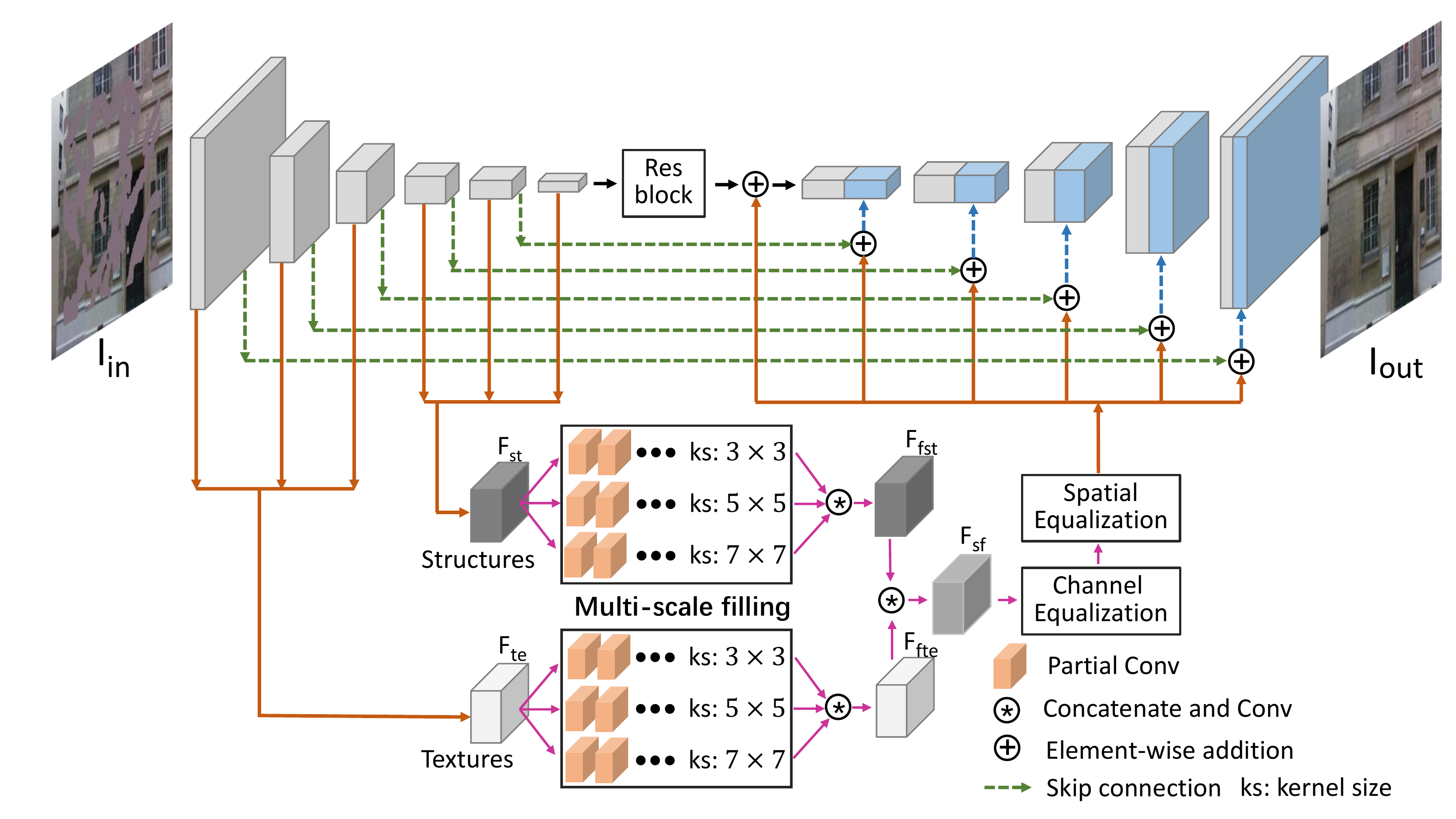}\\
\end{tabular}
\end{center}
\vspace{-5mm}
\caption{Overview of the proposed pipeline. We use a mutual encoder-decoder to jointly recover structures and textures during hole filling. The deep layer features of the encoder are reorganized as structure features, while the shallow layer features are reorganized as texture features. We fill holes in multi-scales within the CNN feature space and equalize output features in both channel and spatial domains. The equalized features contain consistent structure and texture features at different CNN feature level, and supplement the decoder via skip connections for output image generation.}
\label{fig:pipeline}
\end{figure}

Fig. \ref{fig:pipeline} shows the pipeline of the proposed method. We use one mutual encoder-decoder to jointly learn structure and texture features and equalize them for consistent representation. The details are presented in the following:

\subsection{Mutual Encoder-Decoder}
We use an encoder-decoder for end-to-end image generation to fill holes. The structure of this encoder-decoder is a simplified generative network~\cite{Isola-cvpr17-imgtoimg11} where there are 6 convolutional layers in the encoder and 5 convolutional layers in the decoder. Meanwhile, 4 residual blocks~\cite{he-2016cvpr-deepresidual62} with dilated convolutions are set between the encoders and decoders. The dilated convolutions~\cite{Iizuka-sig17-lgmatch4,nazeri-2019ICCVW-edgeconnect46} increase the size of the receptive field to perceive encoder features.

In the encoder, we reorganize the CNN features from deep layers as structure features where the semantics reside. Meanwhile, we reorganize the CNN features from shallow layers as texture features to represent image details. We denote the structure features as $F_{st}$ and the texture features as $F_{te}$ as shown in Fig.~\ref{fig:pipeline}. The reorganization process is to resize and transform the CNN feature maps from different convolutional layers to the same size, and concatenate them accordingly.

After CNN feature reorganization, we design two branches (i.e., the structure branch and the texture branch) to separately perform hole filling on $F_{te}$ and $F_{st}$. The architectures of these two branches are the same. In each branch, there are 3 parallel streams to fill holes in multiple scales. Each stream consists of 5 partial convolutions~\cite{liu-2018eccv-particalcov37} with the same kernel size while the kernel size differs among different streams. By using different kernel sizes, we perform multi-scale filling in each branch for the input CNN features. The filled features from 3 streams (i.e., 3 scales) are concatenated and mapped to the same size of the input feature map via a $1\times1$ convolution. We denote the output of the structure branch as $F_{fst}$, and the output of the texture branch as $F_{fte}$. To ensure the hole filling to focus on the textures and structures, we incorporate supervisions on $F_{fst}$ and $F_{fte}$. We use a $1\times1$ convolution to separately map $F_{fst}$ and $F_{fte}$ to a color image $I_{ost}$ and a color image $I_{ote}$, respectively. The pixel-wise L1 loss can be written as follows:
\begin{equation}\label{eq:feat1}
\begin{aligned}
    L_{rst} = \lVert I_{ost}-I_{st} \rVert_1\\
    L_{rte} = \lVert I_{ote}-I_{gt} \rVert_1
\end{aligned}
\end{equation}
where $I_{gt}$ is the ground truth image and $I_{st}$ is the structure image of $I_{gt}$. We use an edge-preserving smoothing method RTV~\cite{xu-2012TOG-structuredelete53} to generate $I_{st}$ following~\cite{ren-2019iccv-structureflow47}.

The hole regions in $F_{te}$ and $F_{st}$ are filled via structure and texture branches, individually. The feature representations in $F_{fte}$ and $F_{fst}$ are not consistent to reflect the recovered structures and textures. This inconsistency leads to blur and artifacts within and around the hole regions as shown in Fig.~\ref{fig:intro}. To mitigate these effects, we concatenate $F_{fte}$ and $F_{fst}$ first, and make a simple fusion to generate $F_{sf}$ via a $1\times1$ convolutional layer. The texture and structure representations in $F_{sf}$ are corrected via feature equalization at different CNN feature levels (i.e., across shallow to deep CNN layers).

\subsection{Feature Equalizations}

We equalize the fused CNN features $F_{sf}$ in both channel and spatial domains. The channel equalization follows the squeeze and excitation operation~\cite{hu-2018CVPR-squeeze55} to ensure that the attentions within each channel of $F_{sf}$ are the same. As the reweighed channels are influenced by both structure and texture representations in $F_{sf}$, the consistent attentions indicate that these representations are set to be consistent as well. We propagate channel equalization to the spatial domain via the proposed bilateral propagation activation function (BPA).

\subsubsection{Formulation.}

BPA is inspired by the edge-preserving image smoothing~\cite{tomasi-1998CVPR-bilateral57} to generate response values based on spatial and range distances.
It can be written as follows:
\begin{alignat}{2}
 y^s_i&=\frac{1}{C(x)} \sum_{j\in s} g_{\alpha_s} (\lVert j-i \rVert) x_j \label{eqspa}\\
y^r_i&=\frac{1}{C(x)} \sum_{j\in v}  f(x_i,x_j) x_j \label{eqran}\\
 y_i&=q (y^s_i, y^r_i) \label{eqfina}
\end{alignat}
where $x_i$ is the feature channel at position $i$ of input feature $x$, $x_j$ is a neighboring feature channel around $i$ at position $j$, $y^s_i$ and $y^r_i$ are the feature channels after spatial and range similarity measurements. We set the normalization factor as $C(x) = N$, where $N$ is the number of positions in $x$. We use $q$ to denote the concatenation and channel reduction of $y^s_i$ and $y^r_i$ via a $1\times1$ convolutional layer.

The bilateral propagation utilizes the distances of feature channels from both spatial and range domains. We explore $j$ within a neighboring region $s$, which is set as the same spatial size of the input feature for global propagation. The spatial contributions from neighboring feature channels are adjusted via a Gaussian function $g_{\alpha_s}$. When computing $y_i^r$, we measure the similarities between feature channels $x_i$ and $x_j$ via $f(.)$ within a neighboring region $v$ around $i$. The size of $v$ is $3\times3$.
To this end, the bilateral propagation considers both global continuity via $y^i_s$ and local consistency via $y^i_r$.

During the range similarity computation step, we define the pairwise function $f(.)$ as a dot product operation, which can be written as follows:
\begin{equation}\label{eqf}
f(x_i,x_j)= (x_i)^T(x_j).
\end{equation}



The proposed bilateral propagation shares similarity to the non-local block~\cite{wang-2018CVPR-NONE56} that for each $i$, $\frac{1}{C(x)}f(x_i,x_j)$  becomes the softmax computation along  dimension $j$. The difference resides on the region design of propagation. The non-local block uses feature channels from all the positions to generate $y_i$ and the similarity is only measured between $x_i$ and $x_j$. In contrast, BPA considers both feature channel similarity and spatial distance between $x_i$ and $x_j$ during bilateral weight computation. In addition, we use a global region $s$ to compute spatial distance while using a local region $v$ to compute range distance. The advantage of global and local region selections is that we ensure both long-term continuity in the whole spatial region and local consistency around the current feature channel. The boundaries of hole regions are unified with the neighboring image content and the content within the hole regions are set to be consistent.


\begin{figure}[t]
\centering
\includegraphics[width=0.9\linewidth]{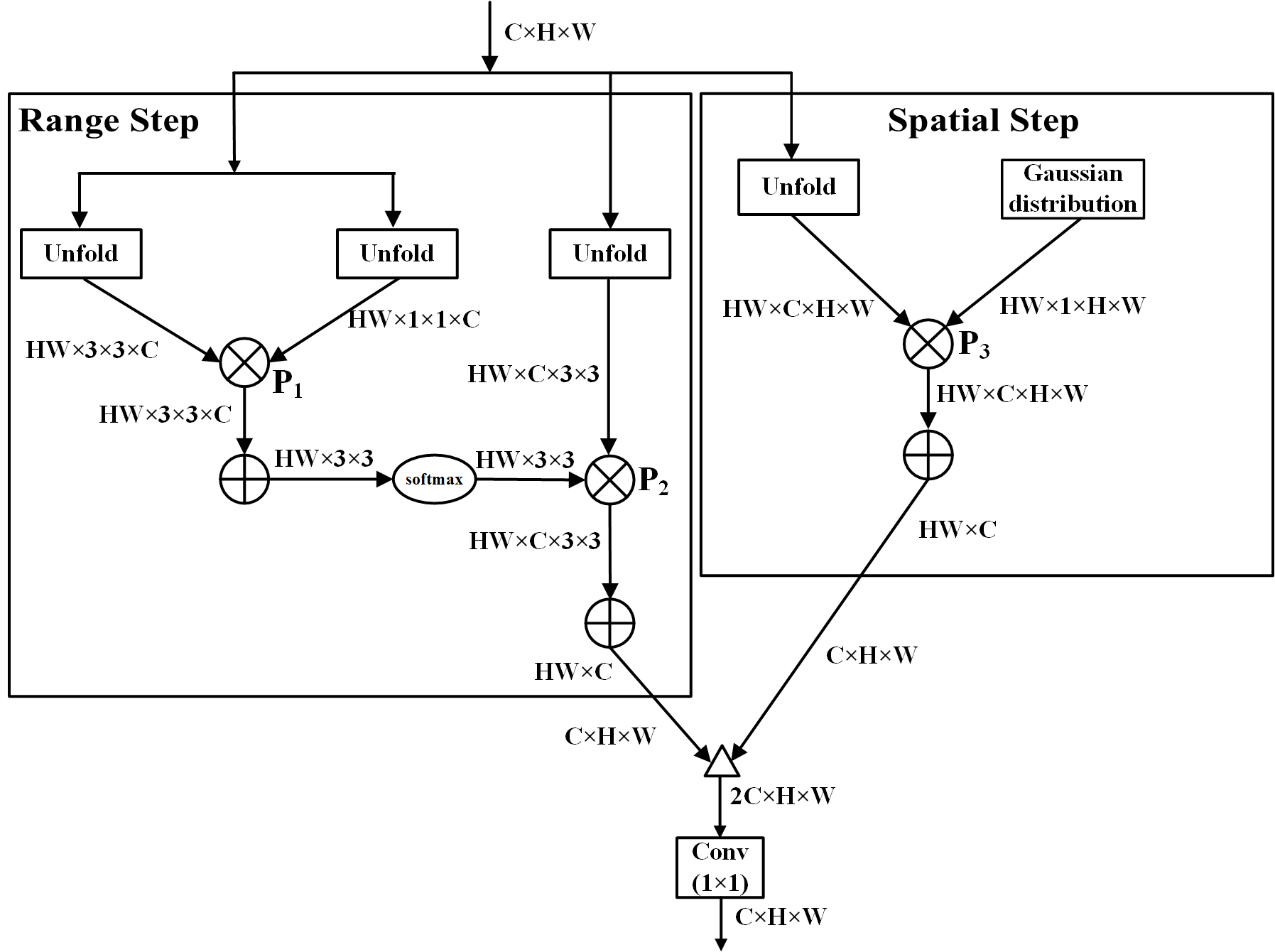}
\vspace{-2mm}
\caption{The pipeline of the bilateral propagation activation function. We denote the broadcast dot product operation as $\otimes$, element-wise addition in selected channel as $\oplus$, and the concatenation as $\bigtriangleup$. For two matrices with different dimensions, broadcast operations first broadcast features in each dimension to match the dimensions of the two matrices.}
\label{fig:BPA}
\end{figure}

\subsubsection{Implementations.}
Fig.~\ref{fig:BPA} shows how bilateral propagation operates in the network. The range step corresponds to the computation of $y_i^r$ in eq.~\ref{eqran} and the spatial step corresponds to $y_i^s$ in eq.~\ref{eqspa}. During range computation, the operations until the element-wise multiplication P$_1$ represent eq.~\ref{eqf} at all spatial locations. We use the unfold function in PyTorch to reshape the feature to vectors (i.e., $HW\times3\times3\times C$) for obtaining all the neighboring $x_j$ for each $x_i$, so that we can make efficient element-wise matrix multiplications. Similarly, the operations until P$_2$ represent the term $\sum_j f(x_i,x_j)\cdot x_j$ in eq.~\ref{eqran}. During spatial computation, the operations until P$_3$ represent the term $\sum_{j} g_{\alpha_s} (\lVert j-i \rVert) x_j$. As a result, the bilateral propagation operation can be efficiently executed via the element-wise matrix multiplications and additions shown in Fig.~\ref{fig:BPA}.

\subsection{Loss functions}
We introduce several loss functions to measure structure and texture differences including pixel reconstruction loss, perceptual loss, style loss and relativistic average LS adversarial loss~\cite{Alexia-ICLR19-rslgan12} during training. We also employ a discriminator with local and global operations to ensure local-global contents consistency, the spectral normalization~\cite{miyato-2018-spectral63} is applied in both local and global discriminator to stable training.

{\flushleft \bf Pixel Reconstruction Loss.} We measure the pixel wise difference from two aspects. The first one is the loss terms illustrated in eq.~\ref{eq:feat1} where we add supervisions on the texture and structure branches. The second one measures the similarity between the network output and the ground truth, which can be written as follows:
\begin{equation}\label{eq:O2 Reconstruct Loss}
    L_{re} = \lVert I_{out}-I_{gt} \rVert_1.
\end{equation}
where $I_{out}$ is the finally predicted image by the network.

{\flushleft \bf Perceptual Loss.} To capture the high-level semantics and simulate human perception of images quality, we utilize the perceptual loss~\cite{Justin-eccv16-perceptual15} $L_{perc}$ defined on the ImageNet-pretrained VGG-16 feature backbone.
\begin{equation}\label{eq:prec}
\begin{aligned}
L_{prec}=\mathbb{E}\Big \lbrack \sum_i \frac{1}{N_i} \lVert\Phi_{i}(I_{out})-\Phi_{i} (I_{gt}) \rVert_1 \Big \rbrack
\end{aligned}
\end{equation}
where $ \Phi_{i}$ is the activation map of the $i$-th layer of VGG-16 backbone. In our work, $\Phi_{i}$ corresponds to the activation maps from layers ReLu1\_1, ReLu2\_1, ReLu3\_1, ReLu4\_1, and ReLu5\_1.

{\flushleft \bf Style Loss.} The transposed convolutional layers from the decoder will bring artifacts that resemble checkerboard. To mitigate this effect, we introduce the style loss. Given feature maps of size $C_j \times H_j \times W_j$, we compute the style loss as follows:
 \begin{equation}\label{eq:style}
\begin{aligned}
L_{style}=\mathbb{E}_j \Big \lbrack  \lVert G_j^\Phi (I_{out})- G_j^\Phi (I_{gt}) \rVert_1 \Big \rbrack
\end{aligned}
\end{equation}
Where $ G_j^\Phi$ is a $C_j \times C_j$  Gram matrix constructed from the selected activation maps. These activation maps are the same as those used in the perceptual loss.

{\flushleft \bf Relativistic Average LS Adversarial Loss.}
We follow~\cite{Yu-cvpr18-attentioninpainting8} to utilize global and local discriminators for perception enhancement. The relativistic average LS adversarial loss is adopted for our discriminators. For the generator, the adversarial loss is defined as:
\begin{equation}\label{eq:LR}
\begin{aligned}
L_{adv}=-\mathbb{E}_{x_{r}}[\operatorname{log}(1-D_{ra}(x_r,x_f))]-\mathbb{E}_{x_{f}}[\operatorname{log}(D_{ra}(x_f,x_r))]
\end{aligned}
\end{equation}
where $D_{ra}(x_r,x_f)=\operatorname{sigmoid}(C(x_r)-\mathbb{E}_{x_{f}}[C(x_f)])$ and $C(.)$ indicates the local or global discriminator without the last sigmoid function. To this end, real and fake data pairs $(x_r,x_f)$ are sampled from the ground-truth and output images.

{\flushleft \bf Total Losses.}
The whole objective function of the proposed network can be written as:
\begin{equation}\label{eq:final loss}
\begin{gathered}
L_{total} = \lambda_r L_{re}+ \lambda_pL_{prec}+\lambda_sL_{style}+\lambda_{adv}L_{adv}+\lambda_{st}L_{rst}+\lambda_{te}L_{rte}
\end{gathered}
\end{equation}
where $\lambda_r$, $\lambda_p$, $\lambda_s$, $\lambda_{adv}$, $\lambda_{st}$ and $\lambda_{te}$ are the tradeoff parameters. In our implementation, we empirically set $\lambda_r=1$, $\lambda_p=0.1$, $\lambda_s=250$, $\lambda_{adv}=0.2$, $\lambda_{st}=1$, $\lambda_{te}=1$.

\def\swfour{0.22\linewidth}
\begin{figure}[t]
\centering
\begin{tabular}{cccc}
\includegraphics[width=\swfour]{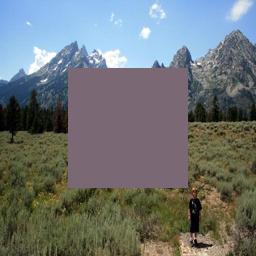}&
\includegraphics[width=\swfour]{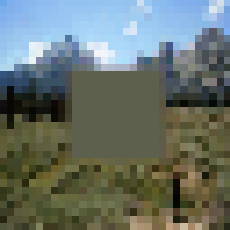}&
\includegraphics[width=\swfour]{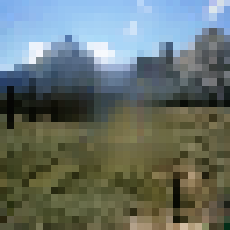}&
\includegraphics[width=\swfour]{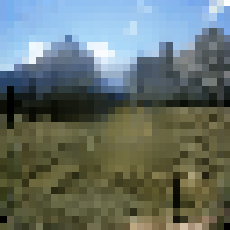}\\
(a) Input &(b) $F_{te}$ &(c) $F_{fte}$ &(d) $F_{sf}$ \\
\includegraphics[width=\swfour]{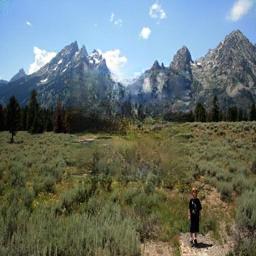}&
\includegraphics[width=\swfour]{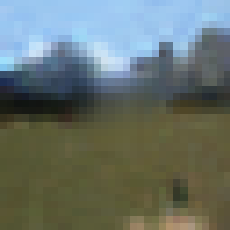}&
\includegraphics[width=\swfour]{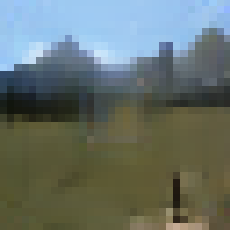}&
\includegraphics[width=\swfour]{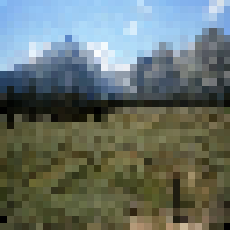}\\
(e) Output &(f) $F_{st}$ &(g) $F_{fst}$ &(h) $F_{equal}$  \\
\end{tabular}
\vspace{-2mm}
\caption{Visualization of the feature map response. The input and output images are shown in (a) and (e), respectively. We use a $1\times 1$ convolutional layer to map high dimension feature maps to the color images as shown in (b)-(d) and (f)-(h).}
\label{fig:visual}
\end{figure}

\subsection{Visualizations}

We use a structure branch and a texture branch to separately fill holes in CNN feature space. Then, we perform feature equalization to enable consistent feature representations in different feature levels for output image reconstruction. In this section, we visualize the feature maps during different steps to show whether they correspond to our objectives. We use a $1\times 1$ convolutional layer to map CNN feature maps to color images for a clear display.

Fig.~\ref{fig:visual} shows the visualization results. The input image is shown in (a) with a mask in the center. The visualized $F_{te}$ and $F_{st}$ are shown in (b) and (f), respectively. We observe that textures are preserved in (b) while the structures are in (f). By multi-scale hole filling, the hole regions in $F_{fte}$ and $F_{fst}$ are effectively reduced as shown in (c) and (g). After equalization, the hole regions in (h) are effectively filled and the equalized features contribute to the decoders to generate the output image as shown in (e).

\renewcommand{\tabcolsep}{0.5pt}
\def\swtwo{0.16\linewidth}
\begin{figure*}[!t]
\centering
\begin{tabular}{cccccc}
\vspace{-0.5mm}
\includegraphics[width=\swtwo]{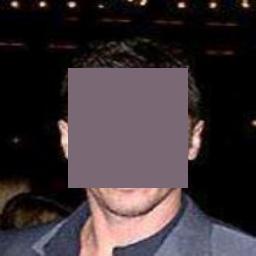}&
\includegraphics[width=\swtwo]{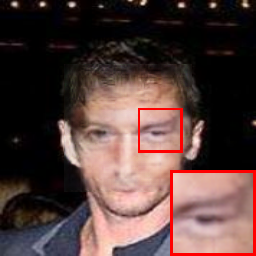}&
\includegraphics[width=\swtwo]{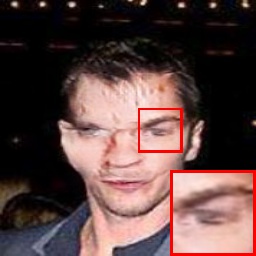}&
\includegraphics[width=\swtwo]{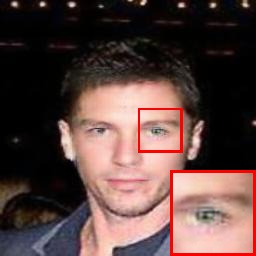}&
\includegraphics[width=\swtwo]{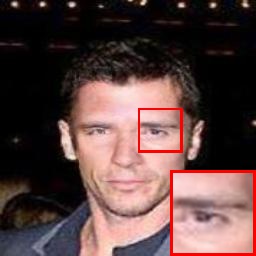}&
\includegraphics[width=\swtwo]{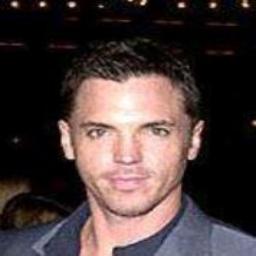}\\
\includegraphics[width=\swtwo]{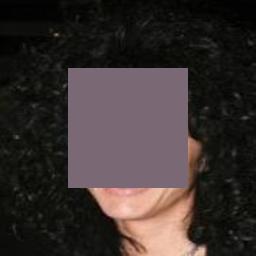}&
\includegraphics[width=\swtwo]{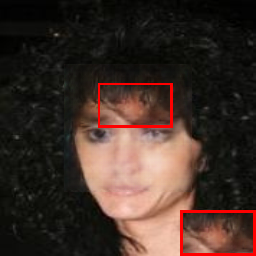}&
\includegraphics[width=\swtwo]{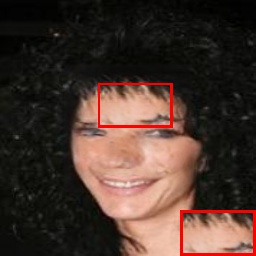}&
\includegraphics[width=\swtwo]{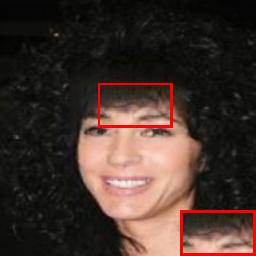}&
\includegraphics[width=\swtwo]{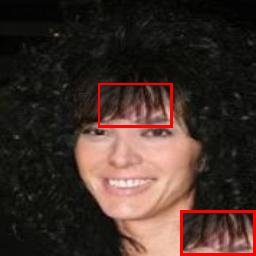}&
\includegraphics[width=\swtwo]{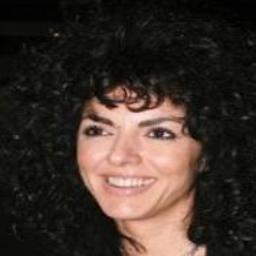}\\
(a) Input &(b) CE~\cite{Pathak-cvpr16-featinpaint7} &(c) CA~\cite{Yu-cvpr18-attentioninpainting8}&(d)  SH~\cite{Yan-eccv18-shiftnet9}&(e) Ours&(f) GT
\end{tabular}
\caption{Visual evaluations for filling center holes. Our method performs favorably against existing approaches to retain both structures and textures.}
\label{imgcenter}
\end{figure*}

\section{Experiments}

We evaluate our method on three datasets: Paris StreetView~\cite{doersch-2015acm-makesparis39}, Place2~\cite{Zhou-TPAMI17-place2su14} and CelebA~\cite{Liu-iccv15-faceat13}. We follow the training, testing, and validation splits of these three datasets. Data augmentation such as flipping is also adopted during training. Our model is optimized by the Adam optimizer~\cite{kingma-2014-adam41} with a learning rate of $2\times10^{-4}$ on a single NVIDIA 2080TI GPU. The training of CelebA model, Paris StreetView model and Place2 model are stopped after 6 epochs, 30 epochs and 60 epochs, respectively. All the masks and images for training and testing are with the size of 256$\times$256.

We compare our method with six state-of-the-art method: CE~\cite{Pathak-cvpr16-featinpaint7}, CA~\cite{Yu-cvpr18-attentioninpainting8}, SH~\cite{Yan-eccv18-shiftnet9}, CSA~\cite{liu-2019iccv-coherent44}, SF~\cite{ren-2019iccv-structureflow47} and GC~\cite{yu-2019iccv-freeform38}. For a fair evaluation on model generalization abilities, we conduct experiments on filling center holes and irregular holes on the input images. The center hole is brought by a mask that covers the image center with a size of $128\times128$. We obtain irregular masks from  PConv~\cite{liu-2018eccv-particalcov37}. These masks are in different categories according to the ratios of the hole regions versus the entire image size (i.e., below 10\%, from 10\% to 20\%, etc). For holes in the image center, we compare with CA~\cite{Yu-cvpr18-attentioninpainting8}, SH~\cite{Yan-eccv18-shiftnet9} and CE~\cite{Pathak-cvpr16-featinpaint7} on the CelebA~\cite{Liu-iccv15-faceat13} validation set. We choose these three methods because they are more effective to fill holes in the image center than fill irregular holes. When handling irregular holes on the input images, we compare with CSA~\cite{liu-2019iccv-coherent44}, SF~\cite{ren-2019iccv-structureflow47} and GC~\cite{yu-2019iccv-freeform38} using Paris StreetView~\cite{doersch-2015acm-makesparis39} and Place2~\cite{Zhou-TPAMI17-place2su14} validation datasets.

\begin{figure*}[!t]
\centering
\begin{tabular}{cccccc}
\includegraphics[width=\swtwo]{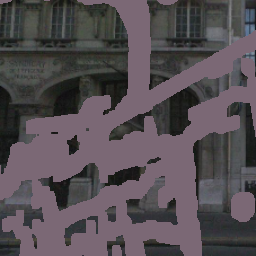}&
\includegraphics[width=\swtwo]{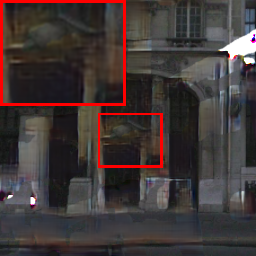}&
\includegraphics[width=\swtwo]{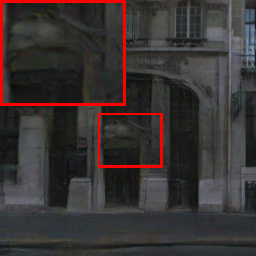}&
\includegraphics[width=\swtwo]{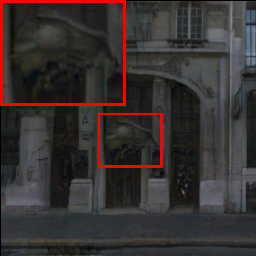}&
\includegraphics[width=\swtwo]{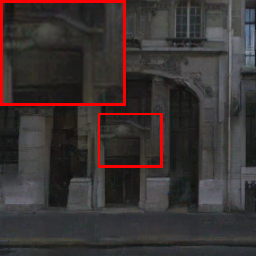}&
\vspace{-0.5mm}
\includegraphics[width=\swtwo]{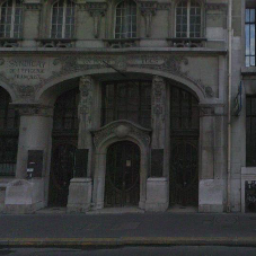}\\
\includegraphics[width=\swtwo]{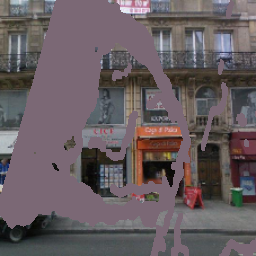}&
\includegraphics[width=\swtwo]{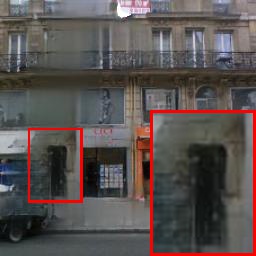}&
\includegraphics[width=\swtwo]{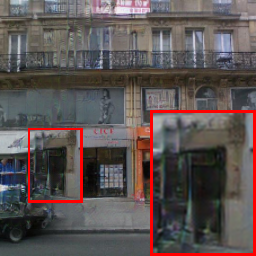}&
\includegraphics[width=\swtwo]{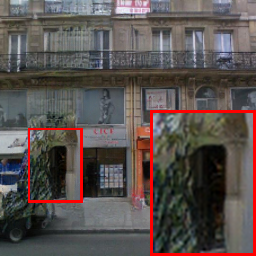}&
\includegraphics[width=\swtwo]{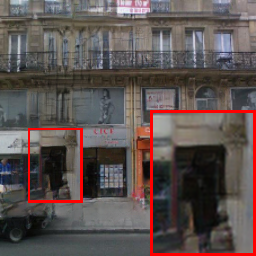}&
\vspace{-0.5mm}
\includegraphics[width=\swtwo]{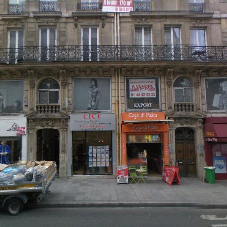}\\
\includegraphics[width=\swtwo]{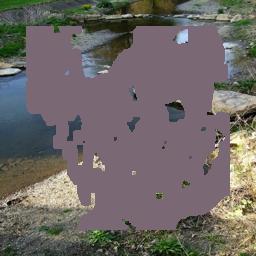}&
\includegraphics[width=\swtwo]{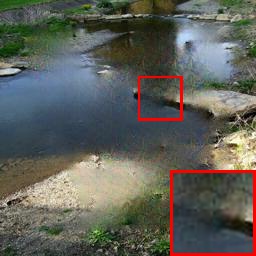}&
\includegraphics[width=\swtwo]{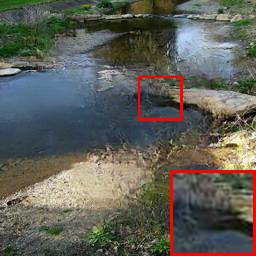}&
\includegraphics[width=\swtwo]{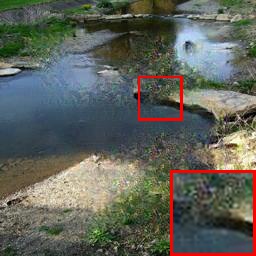}&
\includegraphics[width=\swtwo]{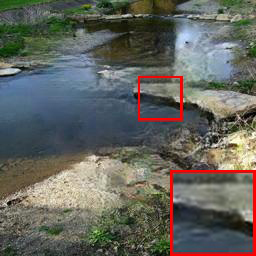}&
\includegraphics[width=\swtwo]{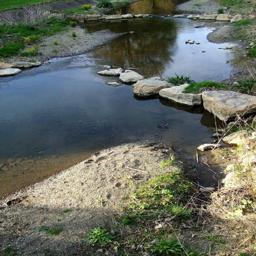}\\
\vspace{-0.5mm}
\includegraphics[width=\swtwo]{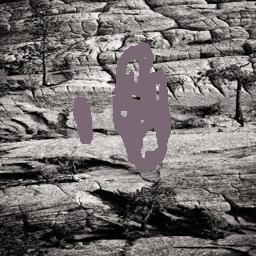}&
\includegraphics[width=\swtwo]{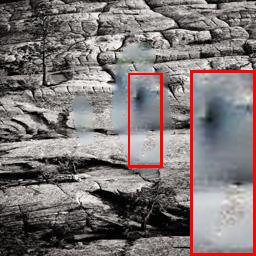}&
\includegraphics[width=\swtwo]{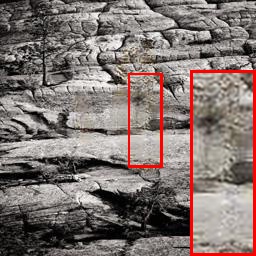}&
\includegraphics[width=\swtwo]{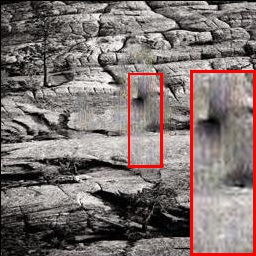}&
\includegraphics[width=\swtwo]{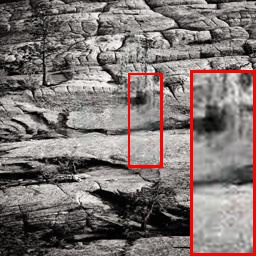}&
\includegraphics[width=\swtwo]{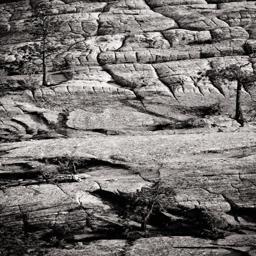}\\
(a) Input &(b) GC~\cite{yu-2019iccv-freeform38}&(c)  SF~\cite{ren-2019iccv-structureflow47}&(d)  CSA~\cite{liu-2019iccv-coherent44} &(e) Ours&(f) GT
\end{tabular}
\caption{Visual evaluations for filling irregular holes. Our method performs favorably against existing approaches to retain both structures and textures.}
\label{imgirre}
\end{figure*}

\subsection{Visual Evaluations}

The visual comparison on the results for filling center holes are in Fig.~\ref{imgcenter} and the results for filling irregular holes are in Fig.~\ref{imgirre}. We also display ground truth images in (f) to show the actual image content. In Fig.~\ref{imgcenter}, the input images are shown in (a). The results produced by CE and CA contains distorted structures and blurry textures as shown (b) and (c). Although more visually pleasing content are generated in (d), the semantics still unreasonable. By utilizing consistent structure and texture features, our method is effective to generate results with realistic textures.

Fig.~\ref{imgirre} shows the comparison for filling irregular holes, which are more challenging than filing centering holes. The results from GC contain noisy patterns shown in (b). The details are missing and the structures are distorted in (c) and (d). These methods are not effective to recover image contents without bringing obvious artifacts (i.e., the second row around the door regions). In contrast, our method learns to represent structures and textures in a consistent formation. The results shown in (e) indicate the effectiveness of our method to produce visually pleasing contents. The evaluations on filling both center holes and irregular holes indicate our method performs favorably against existing hole filling approaches.

\subsection{Numerical Evaluations}

\renewcommand{\tabcolsep}{5pt}
\begin{table}[t]
\centering
 \caption{Numerical evaluations on the CelebA dataset where the inputs are with center hole regions. The $\downarrow$ indicates lower is better while $\uparrow$ indicates higher is better.}
\begin{tabular}{|l|c|c|c|c|c|c|}
\hline
              & CE         & CA       & SH         & Ours                   \\ \hline
FID$\downarrow$       & 52.17     &37.61     &29.72            &\textbf{25.51}                 \\
PSNR$\uparrow$       & 8.53     & 23.65    &26.10             &\textbf{26.32}         \\
SSIM$\uparrow$       & 0.137     &  0.870    &0.902  &\textbf{0.910}                  \\\hline
 \end{tabular}
\label{tab:center}
\end{table}

\renewcommand{\tabcolsep}{3pt}
\begin{table}[t]
\centering
 \caption{Numerical comparisons on the Place2 dataset. The ${\downarrow}$ indicates lower is better while ${\uparrow}$ indicates higher is better.}
 \begin{tabular}{|l|c|c|c|c|c|c|c|}
\hline
         & Mask         & GC          & SF       & CSA           & Ours      \\ \hline
          &10-20\%      & 19.04       & 8.78      &7.85            &\textbf{6.91}   \\
FID$\downarrow$ &20-30\%     & 28.45       & 16.38     & 13.95             &\textbf{8.06} \\
         &30-40\%       & 40.71      & 27.54     & 25.74             &\textbf{19.36}    \\
          &40-50\%      & 60.72        & 40.93     &38.74     &\textbf{28.79}    \\\hline

          &10-20\%      & 27.10     & 29.50   &\textbf{31.31}             &31.13   \\
PSNR$\uparrow$   &20-30\%      & 25.18      & 27.22   & 28.66               &\textbf{28.87} \\
         &30-40\%       & 22.51     &24.37    & 25.01              &\textbf{25.34}    \\
          &40-50\%      & 20.35      & 21.90     & 22.54                 &\textbf{22.81}     \\\hline

          &10-20\%      & 0.929     &0.926    &0.954           &\textbf{0.957}    \\
SSIM$\uparrow$  &20-30\%      & 0.878     & 0.885     & 0.918             &\textbf{0.923}    \\
         &30-40\%       & 0.823    & 0.802     & 0.843              &\textbf{0.854}       \\
          &40-50\%      & 0.670    & 0.678      & 0.702             &\textbf{0.719}      \\\hline
 \end{tabular}
\label{tab:irre}
\end{table}

We conduct numerical evaluations on the Place2 dataset with different mask ratios. Besides, we evaluate numerically on CelebA dataset with center holes in the input images. There are 100 validation images from the ``valley'' scene category chosen for evaluations. In CelebA, we randomly choose 500 images for evaluation. For the evaluation metrics, we follow~\cite{ren-2019iccv-structureflow47} to use  SSIM~\cite{wang-2004tip-ssim64} and PSNR. Moreover, we introduce
FID (F$\rm \grave{r}$echet Inception Distance) metric~\cite{heusel-201nips-fid65} as it indicates the perceptual quality of the results. The evaluation results are shown in Table~\ref{tab:center} and Table~\ref{tab:irre}. Our method outperforms existing methods to fill centering holes. Meanwhile, favorable performance is achieved in our method to fill irregular holes under various hole versus image ratios.

\begin{table}[!t]
\centering
 \caption{Human Subject Evaluation results. Each subject selects the most realistic result without knowing hole regions in advance.}
 \begin{tabular}{|l|c|c|c|c|c|c|c|c|}
\hline
                  & CE         & CA       & SH         & GC   & SF & CSA & Ours  \\ \hline
Paris StreetView  & N/A     &N/A     &N/A              &5.3\%   &21.0\%  &29.8\%    &\textbf{43.7\%}              \\
Place2       &  N/A      &  N/A     & N/A             &3.0\%  &25.0\%   &29.6\%   &\textbf{42.4\%}           \\
CelebA       & 1.2\%      & 2.0\%      &40.4\%      & N/A  & N/A     & N/A    &\textbf{56.4\%}   \\\hline

\end{tabular}
\label{tab:user}
\end{table}

{\flushleft \bf Human Subject Evaluation.} We follow~\cite{zeng-2019cvpr-pyramidfilling66} to involve over 35 volunteers for evaluating the results on CelebA, Place2 and Paris StreetView datasets. The volunteers are all image experts with image processing background. There are 20 questions for each subject. In each question, the subject needs to select the most realistic result from 4 results generated by different methods without knowing the hole region in advance. We tally the votes and show the statistics in Table~\ref{tab:user}. Our method performs favorably against existing methods.

\renewcommand{\tabcolsep}{1pt}
\begin{table}[!t]
\centering
 \caption{Ablation study on the Paris StreetView dataset. Structure and textures improve our performance.}
\begin{tabular}{|c|c|c|c|}
\hline
     & \begin{tabular}[c]{@{}c@{}}Ours without\\ textures\end{tabular} & \begin{tabular}[c]{@{}c@{}}Ours without\\ structures\end{tabular} & Ours           \\ \hline
FID${\downarrow} $  & 30.37                                                           & 27.46                                                             & \textbf{25.10} \\
PSNR${\uparrow} $  & 22.80                                                           & 22.96                                                             & \textbf{23.38} \\
SSIM${\uparrow} $  & 0.818                                                           & 0.823                                                             & \textbf{0.833} \\ \hline
\end{tabular}
\label{tab:stte}
\end{table}

\renewcommand{\tabcolsep}{1pt}
\begin{table}[!t]
\centering
\caption{Ablation study on the Place2 dataset. Non-local aggregation improves our baseline while feature equalization makes further improvement.}
\begin{tabular}{|c|c|c|c|}
\hline
     & \begin{tabular}[c]{@{}c@{}}Ours without \\ equalization\end{tabular} & \begin{tabular}[c]{@{}c@{}}Non-local\\ aggregation\end{tabular} & Ours           \\ \hline
FID${\downarrow} $  & 29.11                                                                & 24.07                                                           & \textbf{21.26} \\
PSNR${\uparrow} $ & 23.14                                                                & 23.64                                                           & \textbf{24.57} \\
SSIM${\uparrow} $ & 0.837                                                                & 0.848                                                           & \textbf{0.852} \\ \hline
\end{tabular}
\label{tab:feature}
\end{table}

\section{Ablation Study}

\begin{figure}[!t]
\begin{center}
\begin{tabular}{ccccc}
\includegraphics[width=\swfive]{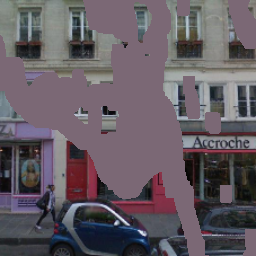}&
\includegraphics[width=\swfive]{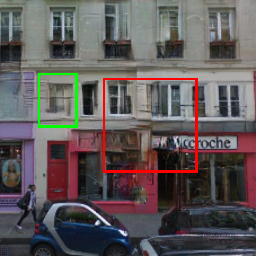}&
\includegraphics[width=\swfive]{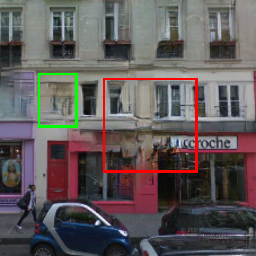}&
\includegraphics[width=\swfive]{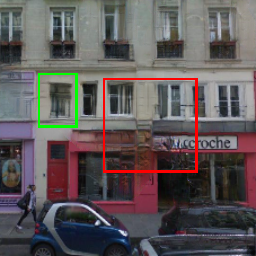}&
\includegraphics[width=\swfive]{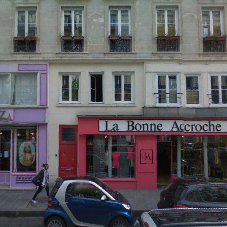}\\
(a) Input & (b) Ours w/o & (c) Ours w/o & (d) Ours & (e) Ground\\
image & textures & structures & & truth\\
\end{tabular}
\end{center}
\vspace{-2mm}
\caption{Abalation studies on structure and texture branches. A joint utilization of these two branches improve the content quality.}
\label{fig:stte}
\begin{center}
\begin{tabular}{ccccc}
\includegraphics[width=\swfive]{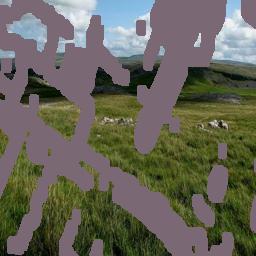}&
\includegraphics[width=\swfive]{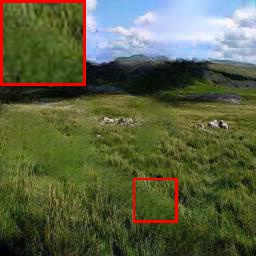}&
\includegraphics[width=\swfive]{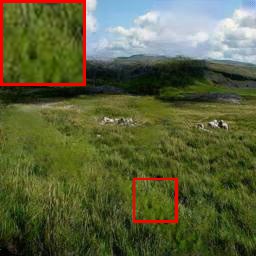}&
\includegraphics[width=\swfive]{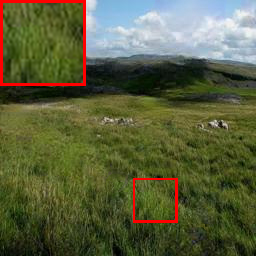}&
\includegraphics[width=\swfive]{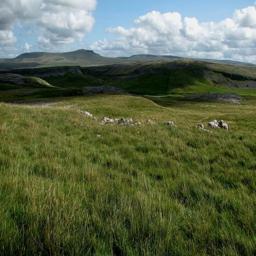}\\
(a) Input & (b) Ours w/o & (c) Non-Local & (d) Ours & (e) Ground\\
image & equalization & aggregation & & truth\\
\end{tabular}
\end{center}
\vspace{-2mm}
\caption{Ablation studies on feature equalizations. More realistic and visually pleasing contents are generated via feature equalizations.}
\label{fig:feq}
\end{figure}

{\flushleft \bf Structure and Texture branches.}
To evaluate the effects of structure and texture branches, we use each of these branches separately for network training. For fair comparisons, we expand the channel number of the texture and structure branch outputs via additional convolutions. So the single branch output contains the same size as that of $F_{sf}$. As shown in Fig.~\ref{fig:stte}, the output of our method without texture branch contains rich structure information (i.e., the window in the red and green boxes) while the textures are missing. In comparison, the output of our method without structure branch does not contain meaningful structure (i.e., the window in the red and green boxes). By utilizing both branches, our method achieve favorable results on both structures and textures. Table~\ref{tab:stte} shows the similar numerical performance on Paris StreetView dataset where these two branches improve our method significantly.

{\flushleft \bf Feature Equalizations.}
We show the contributions of feature equalizations by removing them from the pipeline and showing the performance degradation. Moreover, we show the bilateral propagation activation function (BPA) is more effective to fill hole regions than the Non-local attentions~\cite{wang-2018CVPR-NONE56}. As shown in Fig.~\ref{fig:feq}, without using equalization our method generates visually unpleasant contents and visible artifacts. In comparison, the contents generated by \cite{wang-2018CVPR-NONE56} are more natural. However, the recovered contents are still blurry and inconsistent because the Non-local block ignores the local coherency and global distance of features. This limitation is effectively solved via our method with feature equalizations. Similar performance has been shown numerically on Table~\ref{tab:feature} where our method achieves favorable results.

\section{Concluding Remarks}
We propose a mutual encoder-decoder with feature equalizations to correlate filled structures with textures during image inpainting. The shallow and deep layer features are reorganized as texture and structure features, respectively. In the CNN feature space, we introduce a texture branch and a structure branch to fill holes in multi-scales and fuse the outputs together via feature equalizations. During equalization, we first ensure consistent attentions among each channel and propagate to the whole spatial feature map region via the proposed bilateral propagation activation function. Experiments on the benchmark datasets have shown the effectiveness of our method when compared to state-of-the-art approaches on filling both regular and irregular hole regions.

{\flushleft \bf Acknowledgements.}
This work is partially supported by the National Natural Science Foundation of China under Grant No. 61702176.
%
%
\clearpage
\bibliographystyle{splncs04}
\bibliography{ref}
\end{document}